\RequirePackage{fix-cm}
\documentclass[smallextended]{svjour3}       
\smartqed  
\usepackage{graphicx}
\usepackage[table]{xcolor}
\usepackage{amsmath}
\usepackage{floatrow}
\usepackage{amssymb}
\usepackage{comment}
\usepackage[ruled,vlined,linesnumbered]{algorithm2e}
\usepackage{arevmath}     
\usepackage{xcolor}
\usepackage[position=bottom]{subfig}
\usepackage{textcomp}
\usepackage{xcolor}

\begin{document}

\title{SkiNet: A Deep Learning Solution for Skin Lesion Diagnosis with Uncertainty Estimation and Explainability
}


\author{Rajeev Kumar Singh\ \and Rohan Gorantla \and Sai Giridhar Allada \and Narra Pratap} 
\authorrunning{Rajeev et al.}
%

\institute{Rajeev Kumar Singh\\Shiv Nadar University \\
\email{rajeev.kumar@snu.edu.in}}
\date{Received: date / Accepted: date}

\maketitle

\begin{abstract}
 Skin cancer is considered to be the most common human malignancy. Around 5 million new cases of skin cancer are recorded in the United States annually. Early identification and evaluation of skin lesions is of great clinical significance, but the disproportionate  dermatologist-patient ratio poses significant problem in most developing nations. Therefore a deep learning based architecture, known as SkiNet, is proposed with an objective to provide faster screening solution and assistance to newly trained physicians in the clinical diagnosis process. The main motive behind Skinet's design and development is to provide a white box solution, addressing a critical problem of trust and interpretability which is crucial for the wider adoption of Computer-aided diagnosis systems by the medical practitioners. SkiNet is a two-stage pipeline wherein the lesion segmentation is followed by the lesion classification. In our SkiNet methodology, Monte Carlo dropout and test time augmentation techniques have been employed to estimate epistemic and aleatoric uncertainty, while saliency-based methods are explored to provide post-hoc explanations of the deep learning models. The publicly available dataset, ISIC-2018, is used to perform experimentation and ablation studies.  The results establish the robustness of the model on the traditional benchmarks while addressing the black-box nature of such models to alleviate the skepticism of medical practitioners by incorporating transparency and confidence to the model's prediction. 
\end{abstract}
\keywords{Skin Cancer \and Explainability \and Uncertainty \and Deep Learning \and Medical Imaging}

\section{Introduction}
Skin cancer is the out-of-control growth of abnormal cells in the outermost skin layer known as epidermis \cite{stern2010prevalence}. Every day about 9,500 people are diagnosed with skin cancer in the United States of America (US) \cite{mansouri2017treatment}. There are three main types of skin cancer viz. basal cell carcinoma (BCC), squamous cell carcinoma (SCC), and melanoma. BCC and SCC are the most common forms of skin cancer, with an estimated 4.3 million and 1 million cases reported each year in the US, respectively. BCC and SCC are highly curable, while melanoma is the deadliest form of skin cancer, with 9000 deaths annually in the US \cite{newmelanoma}. Internationally, skin cancer also poses a significant threat to public health, with 20,000 deaths in Europe, 1200 deaths in Australia, and it also accounts for 2-4 \% of all Asian cancers \cite{forsea2012melanoma,gloster2006skin}.

Early diagnosis of skin cancer is a cornerstone to combat the rising mortality as the chances of survival drop from 99\% to 5\% during its progression to the advanced stage \cite{rogers2015incidence}. The diagnosis of skin is primarily conducted through Dermoscopic analysis and specially trained medical practitioners demonstrated clinical accuracy close to 75 \% \cite{ali2012systematic}. There is a dearth of dermatologists globally and in particular, some of the developing nations like Australia and New Zealand suffer from a serious shortage of trained practitioners\cite{arik2017deep}. The research community has made significant efforts to develop automated image analysis systems to detect skin diseases from dermoscopy images in order to overcome the limited supply of experts and provide faster screening solutions. 

The use of Deep Learning (DL) based tools as a diagnostic aid is a growing trend in dermatology. Further, the advent of Convolutional Neural Networks (CNNs) stimulated the research in various aspects of visual recognition tasks which were highly relevant in the context of medical image analysis \cite{singh2020dmenet}. CNNs integrating disease taxonomy were developed to automate the segmentation and classification for skin lesion diagnosis. While these systems improved the accuracy significantly, the faith of doctors on these systems did not witness any major upward trend owing to the black-box nature of such CNN based models. 

The last decade saw rapid progress of DL research in healthcare across various domains with diverse applications however only a few examples of such techniques are successfully deployed into clinical practice. Medical practitioners must be convinced about the efficacy and accuracy of these systems, however, these models need to suffice at least two primary criteria to gain their trust. The model should have the capability to denote the confidence in each prediction and should be interpretable, i.e., it should clearly represent the features that contributed to the prediction\cite{kelly2019key}. 
The model performance is usually presented in terms of metrics related to the discriminative power of the models such as sensitivity, specificity, or ROC curves \cite{leibig2017leveraging}. However, it is important to understand how confident or certain the model is about a prediction, particularly in clinical practice where diagnostic errors have close to zero tolerance and sometimes difficult cases can require expert review. Estimation of uncertainty can be used not only to determine which samples are difficult to classify, thereby requiring more expert analysis, but also to detect samples that deviate from the data used for model training. The network still can make high assurance predictions when the distribution of training and test data varies. This issue of out of distribution (OOD) sample  is an open problem in the domain of DL. Often a crucial problem in the medical setting is deciding whether a model is being used in a environment other than the study. 

The machine learning community has traditionally built models that achieve high classification accuracy on a test set, supposedly derived from the same distribution on which the model is being trained. In reality, however, the data the model is being trained on always vary from the data on which the model is being deployed \cite{quionero2009dataset}.	Patients population vary in demographics and in disease presentation between different locations, and these characteristics change with time. Furthermore, datasets are mostly obtained from a few sites with specific procedures for the acquisition of images that may not generalize to other sites \cite{thiagarajan2018understanding}. For this reason, it is important to comprehend, how a model makes predictions, beyond optimizing performance on a predefined test set. This provides clinicians with insight as to, when the model will fail. Such intuition enables better model development by targeting data collection to challenge out-of-distribution samples, or by modifying model architectures or by using loss functions to reduce these errors. Moreover, when the model makes a prediction for an inappropriate reason, instead of showing the prediction, the system may refer patients to clinicians. Saliency maps have become a common post-hoc interpretability method for CNNs. These maps are designed to highlight the salient components of medical images which are critical for prediction of the model. 

To overcome the aforementioned issues in deploying the skin lesion models in a clinical setting, we incorporate uncertainty estimation and explainability to the proposed SkiNet algorithm with an intention to provide a white-box model for skin lesion diagnosis. Taking inspiration from \cite{gal2016dropout,gal2015bayesian}, we take advantage of the Monte Carlo Dropout technique and Test-time augmentation, which is computationally tractable and provides a fairly close uncertainty estimate of the model. To make the model explainable, we use saliency-based methods given by \cite{springenberg2014striving,selvaraju2017grad,kapishnikov2019xrai}. The main contributions of this article are:

\begin{itemize}
     \item A robust CNN-based CAD model known as SkiNet which incorporates uncertainty estimate and post-hoc interpretation in model predictions.

    \item A novel segmentation algorithm known as Bayesian MultiResUNet is proposed, which provides an uncertainty estimate along with the segmentation map. 
    
    \item Comparative analysis of various saliency methods is performed to understand the optimal technique for post-hoc interpretation of skin lesion diagnosis.

\end{itemize}

This paper is organised as given below. Section \ref{rw} provides a brief overview of the various approaches towards medical image processing and how uncertainity estimation has been used to determine the confidence of a prediction. It also briefly discusses various explainability techniques. The proposed SkiNet pipeline and its various components are discussed in section \ref{methodology}. In section \ref{expmt}, we talk about the datasets and the different metrics used in order to measure the performance of our proposed SkiNet pipeline. Section \ref{resultsss} presents a comprehensive analysis of each stage of the SkiNet pipeline and demonstrates the superior performance of the proposed solution. Finally, a brief conclusion and its future scope, is given in section \ref{conc}. 
\section{Related Work} \label{rw}

Skin cancer is one of the most common cancers occurring worldwide, and early detection is essential and necessary to improve patient outcomes. The earliest computer-aided diagnostic (CAD) system for the diagnosis of skin cancer can be traced back to the late 1980s when researchers used hand-crafted techniques such as border detection, semi-translucence detection, telangiectasia detection, and ulcer as well as crust detection \cite{MOSS198931}. Segmentation and classification are the two most vital tasks that researchers study for the automation of diagnosis and classification of skin lesions into the respective category of cancer or as a precancer. In addition, the work performed so far has been broadly based on the extraction of hand-crafted features, a combination of hand-coded features with Machine Learning algorithms, and usage of  Deep Learning (DL) techniques for feature detection.
The earliest articles pertaining to computer-aided diagnosis used hand-crafted extraction techniques based on the dermoscopy rule of ABCD, where the skin lesions can be characterized based on \textbf{a}symmetry, \textbf{b}order irregularities, \textbf{c}olor distribution, and \textbf{d}ermoscopic structures\cite{nachbar1994abcd,barata2018survey}. Barata \textit{et al.} \cite{barata2013two} employed two different approaches for the detection of melanomas in dermoscopic images, where in the first technique used global methods to classify skin lesions, while the second used local features and the bag-of-features classifier. This study concluded that when used in isolation, color features outperform texture features and that, both approaches together produce excellent results. Codella \textit{et al.} \cite{codella2017deep} used hand-coded features that involved multi-scale local binary color patterns, color histogram, and edge histogram. Recent advancements in DL have led to the wide usage of CNNs for segmentation \cite{ref10.1007/11752912_23,7903636,7899656,lin2017skin} and classification \cite{liao2016deep,7893267,abbas2019dermodeep} for skin lesion diagnosis. However, these aforementioned DL models act as a black box and lack confidence and explainability, which are essential for the success of such models in the medical domain.

Even though a deep learning model is uncertain about a particular prediction, it would still make a definitive prediction, which might be cataclysmic in the medical diagnosis scenario where the human, economic and social cost of error is very high. Displaying a measure of certainty along with the traditional CAD prediction would allow doctors to adapt their trust according to the confidence as shown by the model. This aspect of certainty and confidence was addressed in recent works like \cite{hershkovitch2018model}, where the stochastic active contour segmentation approach was used to produce a large set of plausible segmentations, and then the weighted sum of these segmentations was calculated to find the uncertain margins. Wang \textit{et al.} \cite{wang2019aleatoric} used test time augmentation for measuring uncertainty in the segmentation of MRI scans. Ghahramani. \textit{et al.} \cite{gal2016dropout} proposed the use of dropouts as a Bayesian approximation in order to calculate the model uncertainty. This method was utilized in \cite{mobiny2019risk,wickstrom2020uncertainty} to measure the uncertainty in classification and segmentation tasks in the medical scenario. Unlike performance metrics such as accuracy, sensitivity, etc. interpretability is not entirely quantifiable; however, it is crucial to understand what the model is learning.
Recent works \cite{lee2018robust,wickstrom2020uncertainty} have deployed techniques like Guided Backprop and Grad CAM to highlight the essential features that contribute to the model's prediction.

\begin{figure}[ht] 
    \centering
    \includegraphics[scale=0.5]{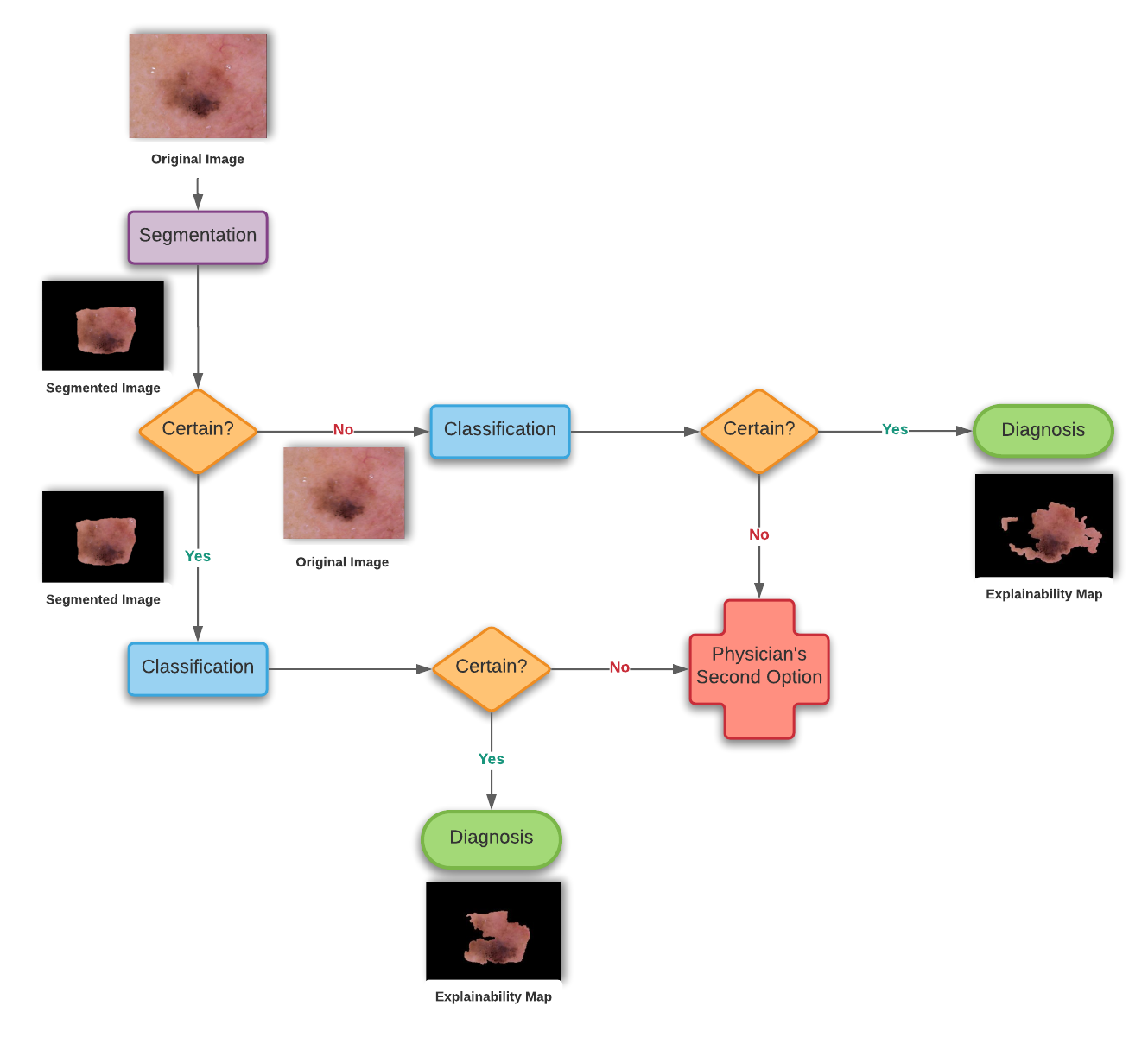}
    \caption{SkiNet pipeline}
    \label{Fig:skinet}
\end{figure}
\section{Proposed Methodology- SkiNet} \label{methodology}

Our proposed SkiNet methodology as illustrated in Fig.\ref{Fig:skinet} is a two-step process which incorporates the uncertainty estimation and the explainability of the algorithm's decision. In the first step, we perform segmentation to extract key regions from the input image and then feed this segmented image to our second step which is classification, provided the segment produced is certain or else the original image itself is passed to the second step. If the proposed algorithm is uncertain about its final diagnosis then it would suggest for expert intervention else it would give results with confidence and also show the key pixels which played an essential role in the decision-making process.
We present our method for estimating the associated uncertainty in section \ref{uncertainity} and then explain how we are incorporating interpretability to our models in section \ref{explainability}. Finally, we discuss our segmentation algorithm in section \ref{seg} and classification strategy in section \ref{class}.

\subsection{Uncertainty Estimation} \label{uncertainity}

CNNs have certain shortcomings despite its progress in a wide range of applications. One of the concerning drawbacks among them is its inability to provide a notion of uncertainty in its prediction, which is crucial in the medical domain \cite{gal2016uncertainty}. For example, the hypothetical model should return a prediction with reasonably high confidence in a scenario where the CNN model trained on a variety of car data, trying to predict the category to which the given car belongs. But what should happen if the model is tested with a image of bike and asks the model to decide on a car category? This is a situation where the test data is far from the distribution as the model is trained on distinguishing among various car classes and have never seen the image of a bike. In such cases, the model is expected to return not only a prediction but also some additional details communicating the high degree of uncertainty with these kinds of data. Uncertainty estimate can be used not only to assess which samples are difficult to identify, thereby requiring a further expert review, but also to detect samples that deviate from the data used to train the model. When the distributions of training and test data vary, the network can still make high confidence predictions. This issue is called as the out of distribution problem. In medical domain, determining when a model is being used on a setting different from the training setting is also a critical issue. 

There are mainly two types of uncertainty viz., Aleatoric and Epistemic uncertainty \cite{der2009aleatory}. Aleatoric uncertainty captures noise inherent in the data and cannot be abated by collecting more data \cite{kendall2017uncertainties}. Epistemic uncertainty, also known as model uncertainty, accounts for variability in the parameters of the model and analyzes what the model is not aware owing to the lack of training data \cite{gal2016dropout}. Epistemic uncertainty is necessary for small datasets with imbalance and safety-critical applications, as epistemic uncertainty is needed to understand examples that vary from training data. \cite{kendall2017uncertainties}. 

Uncertainties are formulated as probability distributions over the model parameters (for epistemic uncertainty) or model inputs (for aleatoric uncertainty). Bayesian statistics have largely inspired most of the work done till now on uncertainty estimation techniques. Bayesian Neural Network (BNN) \cite{neal2012bayesian} is the probabilistic variant of the traditional neural networks and provides a mathematical framework for uncertainty estimation. Most of the earlier works on epistemic uncertainty estimation are based on Bayesian inference. However, in practice, Bayesian inference is computationally expensive; therefore, extensive research has been done in developing various techniques to approximate Bayesian deep networks although they are not scalable for larger convolutional networks \cite{mackay1992practical,neal1993bayesian,mackay1995probable,blundell2015weight}. Research has also been carried out to develop alternative strategies, which are suitable for approximating the uncertainty \cite{gal2016dropout,mobiny2019dropconnect}.  The work proposed by  \cite{gal2016dropout} demonstrated how dropout \cite{srivastava2014dropout} applied on a neural network with an arbitrary number of layers is mathematically equivalent to estimating variational inference in Gaussian process model \cite{damianou2013deep}. This was is later extended to CNNs in \cite{gal2015bayesian} explaining that dropout can be used to enforce a Bernoulli distribution over the weights of the CNN without any additional model parameters. This method is known as Monte Carlo (MC) Dropout and is successfully employed in some of the applications in the medical imaging domain \cite{yang2016fast,leibig2017leveraging}.



The dropout layers are generally added in many deep neural networks to reduce overfitting by randomly dropping weights with a fixed probability. Inspired by the capability of the MC-Dropout technique in estimating uncertainty, we employ the same to build our proposed model. Given a test sample $s^{*}$, we sample the network $O$ times over its parameters and thereby giving an estimate of the predictive posterior distribution. This sampling is known as Monte Carlo sampling and the mean $\mu_e$ over these iterations is considered as the final result on a given test sample. $\mu_e$ is computed as shown in the equation below
\begin{equation}
    \mu_e \approx \frac{1}{O}\sum_{m=1}^{O}p(y^{*}|s^{*},\hat{W}_O)
\end{equation}{}
where $\hat{W}_O$ denotes the weights of the network with dropouts in $O^{th}$ MC iteration and $O$ is the total number of sampled sets of weights. Among several classes $y^{*}$, the one with $\mu_{max}$ is selected as the outcome for each test sample $s*$.

Aleatoric uncertainty captures noise inherent in the data and cannot be abated by collecting more data \cite{kendall2017uncertainties}. Aleatoric uncertainty can be estimated either by learning a mapping directly from the input data \cite{kendall2017uncertainties} or by test time data augmentation \cite{wang2019aleatoric,ayhan2018test,combalia2020uncertainty}. However, the former technique suffers a drawback, as is it involves adapting the network architecture and loss function, which restricts the application to trained models. Therefore, we employ test-time data augmentation technique in our pipeline. In this approach, a test sample $s^{*}$ is augmented to form $V$ different versions of the image and are forwarded to the network. The mean $\mu_a$ over these iterations is considered as the final result on a given test sample. $\mu_a$ is computed as shown in the equation below
\begin{equation}
    \mu_a \approx \frac{1}{V}\sum_{v=1}^{V}p(y^{*}|s^{*}_v,\hat{W})
\end{equation}{}
where $s^{*}_v$ denotes the $v^{th}$ augmented image and $V$ is the total number of image augmentations. Among several classes $y^{*}$, the one with $\mu_{max}$ is selected as the outcome for each test sample $s*$.

These two approaches are then combined to calculate the overall uncertainty where given a test sample $s^{*}$ is augmented to form $M$ different versions of the image and are forwarded to the network with the dropout activated during the test time. The mean $\mu$ over these iterations is considered as the final result on a given test sample. $\mu$ is computed as shown in the equation below
\begin{equation}
    \mu \approx \frac{1}{M}\sum_{m=1}^{M}p(y^{*}|s^{*}_m,\hat{W}_m)
\end{equation}{}
where $s^{*}_m$ denotes the augmented image passed and $\hat{W}_m$ denotes the weights of the network with dropouts during the $m^{th}$ iteration and $M$ is the total number of iterations. Among several classes $y^{*}$, the one with $\mu_{max}$ is selected as the outcome for each test sample $s*$.

In order to estimate the model uncertainty $\varphi$, we calculate the entropy of the averaged probability vector across the $N$ classes.
\begin{equation}
    \varphi = -\sum_{n=1}^{N}p_n\log p_n
\end{equation}{}
where $p_n$ is the probability of $n$th class.

\subsection{Explainability} \label{explainability}
CNNs lack interpretability which is an essential requirement in the medical domain due to the possibility of life threatening consequences. It is important for a medical practitioner to understand the key features in the image, used by the given model, to make predictions so that it can be verified if it is consistent with medical knowledge and also build trust in the model’s capability. While interpretability is desirable in all domains, medical practitioners have to deal with medico legal, ethical, and strict regulations. Recently there has been a considerable amount of research on saliency methods, that relate CNNs prediction to the inputs that has maximum influence on the prediction. These techniques may be useful in a variety of ways, including tracking a model's assessment, ensuring that the model does not learn false correlations, and evaluating the model for issues related to fairness \cite{ribeiro2016should,selvaraju2017grad}. 

Saliency based methods can broadly be classified into two categories. One collection of methods modifies the input and computes the effect of this change on the output by making a forward pass through the network using these altered inputs \cite{fong2017interpretable,carter2018made}. The other set of approaches calculate attributions by returning the prediction score back to the input features through each layer of the network. In general, second category methods are faster than the initial set of methods, as they usually require a single or constant number of neural network queries \cite{kapishnikov2019xrai}. Guided Backprop \cite{springenberg2014striving}, Grad CAM \cite{selvaraju2017grad}, Guided GradCAM \cite{selvaraju2017grad} and XRAI \cite{kapishnikov2019xrai} are some of the promising approaches in this category. Therefore we explore these techniques in our approach to bring model interpretability in the context of skin lesion detection.

\subsubsection{Guided Backpropagation}
Guided Backpropagation is a technique for visualizing CNN by slightly modifying the backpropagation algorithm wherein the negative gradients are set to zero in each layer allowing only positive gradients to flow backward through the network. Guided Backpropagation is a combination of backpropagation and deconvolution. During forward pass, due to the presence of ReLU activations all the negative input values passed through neurons are set to zero. Therefore, during the backward pass of backpropagation the gradients don’t flow back through these neurons. In deconvolution, during the backward pass all the negative gradients are suppressed to zero. In guided backpropagation both the negative gradients and the gradients with negative input are suppressed to zero. The rationale behind this modification is that all the positive gradients of higher magnitude imply key pixels while negative gradients denote the pixels which the model wants to suppress. 


\subsubsection{Grad-CAM}
Grad-CAM provides a visual explanation by leveraging the gradient information coming into the final convolutional layer. The last layer is chosen as it provides the best tradeoff between detailed spatial information and high-level semantics \cite{selvaraju2017grad}. It considers convolutional layer as the convolutional features generally possess the spatial information. The key pixels responsible for categorization into a particular class is determined by first forward propagation through the network by obtaining gradients for each class. The gradients during backpropagation are average-pooled to obtain the weights that are important for the target class prediction. The weights obtained are combined with activations maps using ReLU operation to compute the Grad-CAM heatmap. To generate a Grad CAM heatmap $V^{e}_{Grad-CAM} \in  \mathbb{R}^{w\times h}$ ,of width $w$ and height $h$ for a class of interest $e$ , the gradient of the score of class $e$ , $z^{e}$ is calculated with respect to the feature maps $F^{a}$ i.e$\frac{\partial z^{e}}{\partial F^{a}}$ . Then these gradients are global average pooled to get the neuron importance weights $\beta ^{e}_{a}$: 
\begin{equation}
 \beta ^{e}_{a}= \frac{1}{G} \sum_{k} \sum_{l}  \frac{\partial z^{e}}{\partial F^{a}_{k l}} 
\end{equation}{}
$\frac{1}{G} \sum_{k} \sum_{l}$ correspond to the global average pooling. The $\beta ^{e}_{a}$ represents a partial linearization \cite{selvaraju2017grad} of the network downstream from $F$ for a class of interest $e$.Finally to obtain the heatmap a weighted combination of forward activation maps is computed followed by a ReLU. 
\begin{equation}
 V^{e}_{Grad-CAM} = ReLU\left ( \sum _{a} \beta ^{e}_{a} F^{a} \right )
\end{equation}{}
 
Guided Grad-CAM overcomes the drawback of Grad-CAM which is the inability to show fine-grained importance like Guided Backpropagation which is a pixel-space gradient visualization method  \cite{selvaraju2017grad}. Guided Grad-CAM is the blend of Guided Backpropagation and Grad-CAM algorithms via pointwise multiplication to incorporate the advantages of both the methods. First the heatmap $V^{e}_{Grad-CAM}$ of an input image is obtained via Grad-CAM, then this heatmap is upsampled to the input image resolution finally it is pointwise multiplied with Guided Backpropagation to obtain Guided Grad-CAM visualization. The resultant visualization has high resolution and is class discriminative.

\subsubsection{XRAI}		 	 	 		
XRAI is the most recently proposed saliency method based on Integrated Gradients \cite{sundararajan2017axiomatic} that decides the key inputs by changing the network input from baseline to the original input and consolidating these gradients. It begins by segmenting the image using Felzenswalb’s graph-based segmentation \cite{felzenszwalb2004efficient} technique, followed by repeated testing of the significance of each segment using attributions. Integrated gradients are used as an attribution with black and white baselines to resolve their setback as they are insensitive to pixels similar to or equal to the baseline image. Every pixel gets an equal chance to contribute to the attributions regardless of the distance from the baseline. Finally, it merges regions with a higher positive value of the sum of all the attributions of that region until it has the complete image as the mask or runs out of regions to add \cite{kapishnikov2019xrai}.

\subsection{Segmentation}\label{seg}

\begin{figure}[!ht]
\caption{Bayesian MultiResUNet}\label{BMRU}
\includegraphics[scale=0.12]{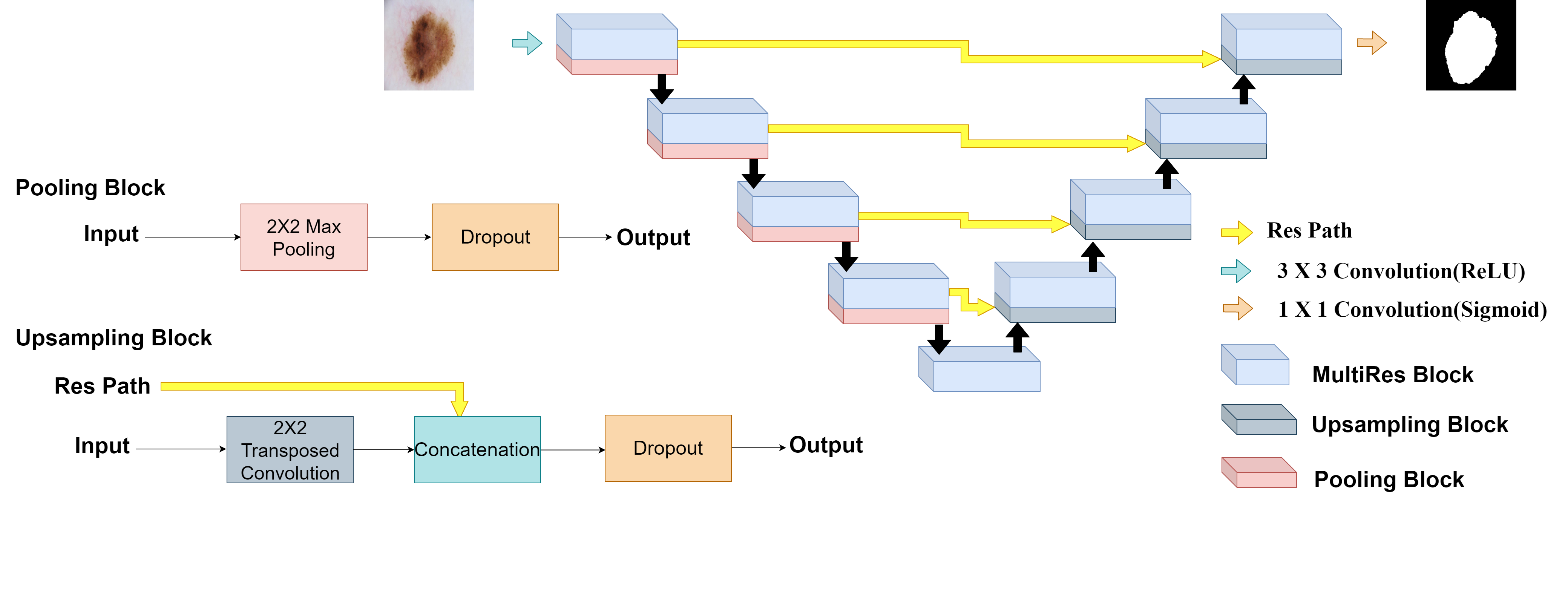}
\end{figure}


In medical image analysis, some pixels in the image contain vital information that might play a crucial role in decision making and thereby providing a rationale for the treatment. Segmentation would help in augmenting the classification model performance in most cases and furthermore, would reduce the computation time \cite{gorantla2019cervical}. In the latter part of the last decade, CNN based segmentation algorithms performed well in biomedical image segmentation tasks. More importantly, U-Net \cite{ronneberger2015u} has been the most promising architecture in this domain and was applied to various segmentation tasks \cite{dalmics2017using,cciccek20163d,poudel2016recurrent}. 

U-Net defined the state of the art in the medical image segmentation tasks \cite{drozdzal2016importance}, however it is not robust enough to analyze objects in the image present at different scales. One of the novel ideas of U-Net architecture has been the implementation of shortcut links between the corresponding layers before the max-pooling and after the deconvolution operations, to relay the spatial information that gets lost from encoder to decoder during the pooling process. The dispelled spatial features though retained still suffers from shortcomings in the skip connections i.e., there is a plausible semantic gap between the two sets of features being merged. The features from the encoder are supposed to be lower-level features, and on the contrary, the decoder features are much higher because they come from deeper layers after fairly complex computation \cite{ibtehaz2020multiresunet}.
\begin{figure}[!ht]
    \centering
    \includegraphics[scale=0.35]{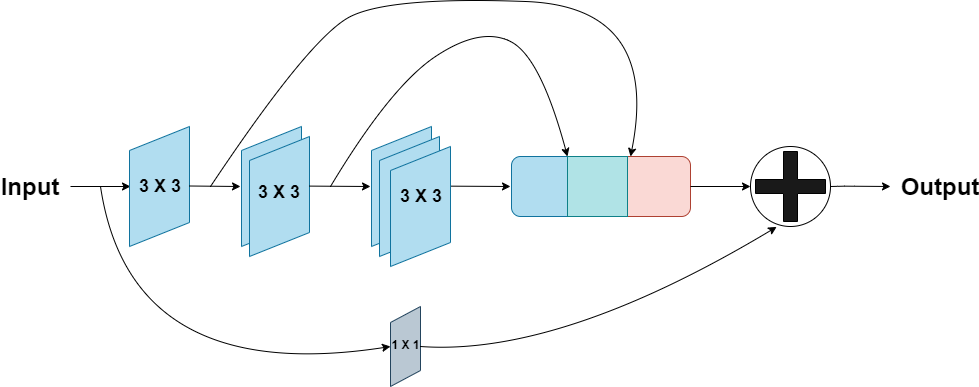} 
    \caption{MuliRes Block}
    \label{Fig:mrb}
\end{figure}
In order to tackle the shortcomings discussed above, \cite{ibtehaz2020multiresunet} proposed few structural changes in the form of \textit{'MultiRes block'} and \textit{'Res path'} to the U-Net architecture drawing inspiration from \cite{drozdzal2016importance,szegedy2016rethinking,szegedy2017inception}. Inspired by the sucessful working of MutiRes block and Res path structures, we employ it in our segmentation architecture known as \textit{Bayesian MultiResUNet}. Similar to the Inception blocks \cite{szegedy2015going}, where convolutional layers of different kernel sizes are adopted to inspect the points of interest in images from different scales, MultiRes blocks employs $3 \times 3$, $5 \times 5$ and $7 \times 7$ filters in parallel with the larger and computationally expensive $5 \times 5$ and $7 \times 7$ blocks factorized as a succession of $3 \times 3$ without affecting the objective function \cite{szegedy2016rethinking}. Additionally, MultiRes blocks contain $1 \times 1$ convolutional layers, for better comprehension of spatial information as shown in Fig. \ref{Fig:mrb}. Rather than just concatenating the feature maps from the encoder stages to the decoder stages as in the shortcut connection of U-Net, Res paths transfers them through a chain of convolution layers with residual connections and then concatenates them with the decoder features to mitigate the gap between encoder and decoder features. Res path is represented in Fig. \ref{Fig:rp} below. 
\begin{figure}[!ht]
    \centering
    \includegraphics[scale=0.3]{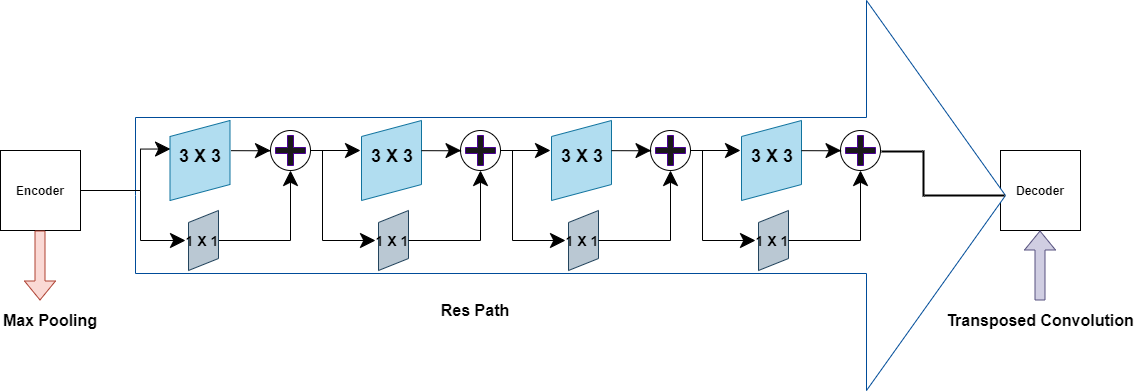} 
    \caption{Res Path}
    \label{Fig:rp}
\end{figure}
As shown in Fig. \ref{BMRU}, Bayesian MultiResUNet has symmetric architecture where the encoder is responsible for extracting spatial features from the input image while the decoder produces the segmentation map using the encoded features. In the encoder, the weights obtained from the MultiRes block are passed to a pooling block where a dropout layer is appended after the pooling operation and these acquired weights are used as an input to the next MultiRes block. The fifth MultiRes block acts as a bridge between encoder and decoder with three $3 \times 3$ convolution operations followed by one $1 \times 1$ convolutional operation. On the other hand, the decoder begins at the upsampling block, which incorporates $2 \times 2$ transposed convolution operation \cite{zeiler2010deconvolutional} to perform upsampling thereby reducing the feature channels by half. These weights are then passed on to the MultiRes block, similar to the encoder. This succession of upsampling and MultiRes operations is repeated four times, reducing the number of filters by two at each stage. Finally, a $1 \times 1$ convolution operation is performed to generate the segmentation map. As we step towards the inner shortcut routes, the intensity of the semantic gap between the encoder and the decoder function maps would possibly decrease; thus we reduce the number of convolutional blocks, i.e., we employ 4, 3, 2, 1 convolutional blocks respectively along the four Res paths. Also, we use 32, 64, 128, 256 filters in the blocks of the four Res paths respectively to compensate for the number of feature maps in encoder-decoder similar to \cite{ibtehaz2020multiresunet}. ReLU \cite{lecun2015deep} activation function and batch-normalization \cite{ioffe2015batch} are employed by all convolutional layers in this architecture, except for the final one which uses a Sigmoid activation function. 
\subsection{Classification} \label{class}
In our experimentation, we considered some of the popular, state-of-the-art deep neural network architectures, including ResNet-50 \cite{he2016deep}, and DenseNet-169 \cite{huang2017densely}. To obtain the Bayesian version of these networks, dropout layers are added as discussed in section \ref{uncertainity}. The dropout can even degrade the performance of the model, therefore we empirically evaluate the performance of several Bayesian models with various configurations which include the positioning of dropout layers as well as the dropout rate, to identify those with the best performance of prediction for the skin lesion classification task. Moreover, all the Bayesian networks employed in our analysis are approximate Bayesian models as the exact Bayesian inference for neural networks is computationally intractable.

\section{Experimentation}\label{expmt}
\subsection{Datasets}\label{datasets}
Task 1 and 3 data of ISIC 2018 have been used for segmentation and classification, respectively \cite{codella2019skin}. 
\subsubsection{Classification}\label{classdata}
 The task 3 data contained 10015 dermoscopic images over 7 classes viz. Melanoma (MEL), Melanocytic Nevi (NV), Basal Cell Carcinoma (BCC), Actinic Keratoses and Intraepithelial Carcinoma (AKIEC), Benign Keratosis (BKL), Dermatofibroma (DF), and Vascular lesions (VASC) as represented in figure 5. The dataset suffers from severe class imbalance issues; hence the data was augmented through vertical, horizontal flipping and random rotations in the range of $[-65,65]$.

\begin{figure}[!ht] 
\subfloat[MEL]{
  \includegraphics[width=1.68cm]{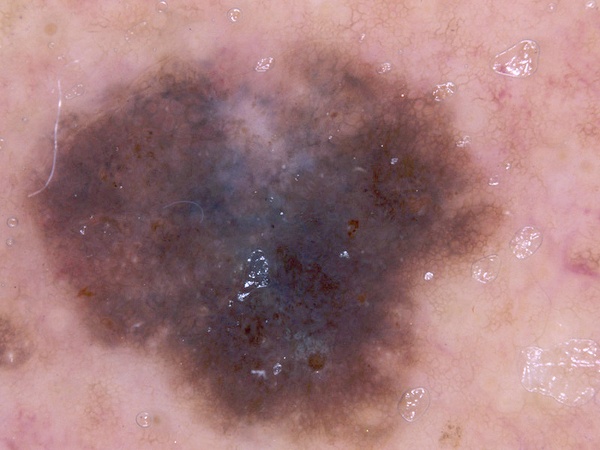}
}
\subfloat[NV]{
  \includegraphics[width=1.68cm]{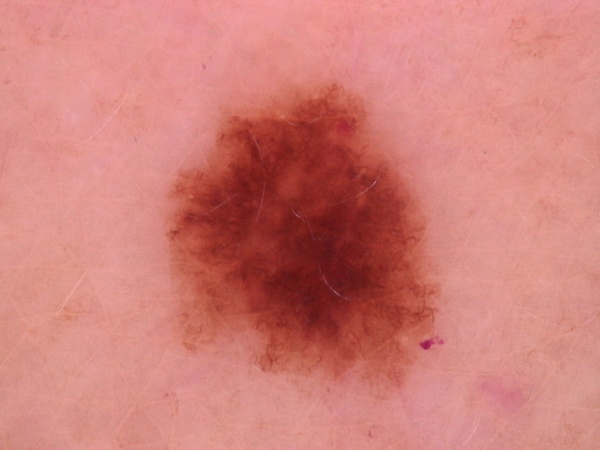}
}
\subfloat[BCC]{
  \includegraphics[width=1.68cm]{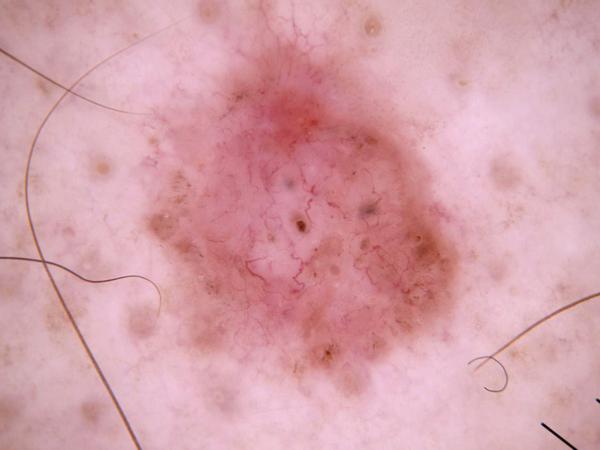}
}\subfloat[AKIEC]{
  \includegraphics[width=1.68cm]{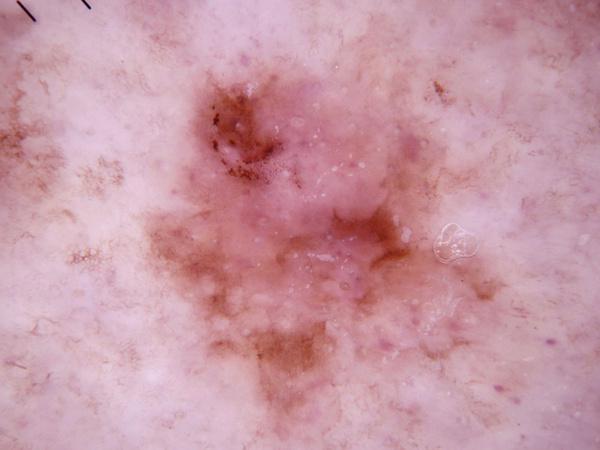}
}
\subfloat[BKL]{
  \includegraphics[width=1.68cm]{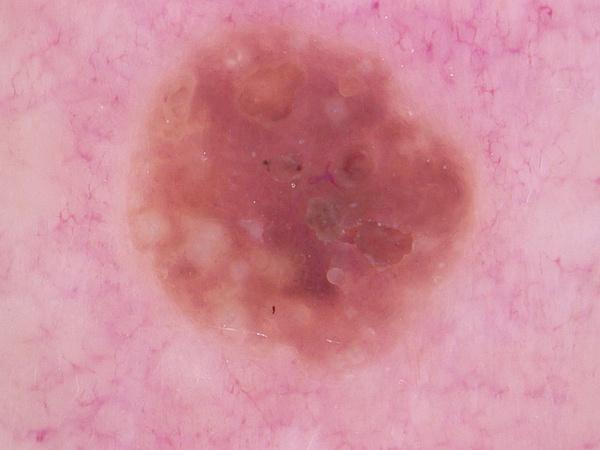}
}
\subfloat[DF]{
  \includegraphics[width=1.68cm]{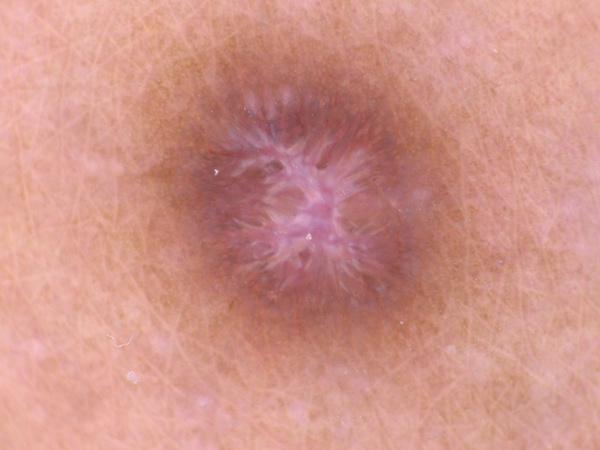}
}
\subfloat[VASC]{
  \includegraphics[width=1.68cm]{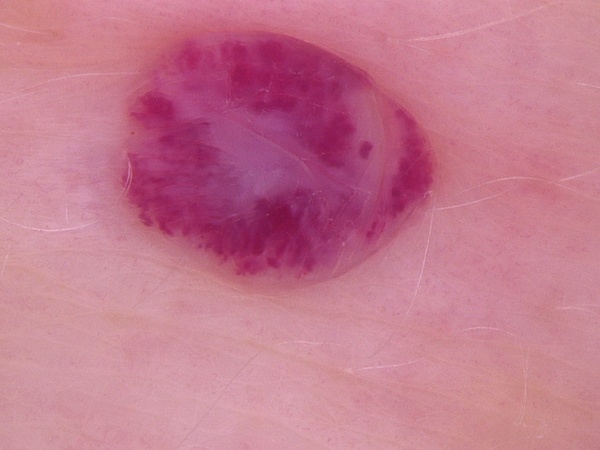}
}

\caption{Visual examples depicting the seven categories of pigmented skin lesions}\label{Fig:all_classi}
\end{figure}
\subsubsection{Segmentation}\label{segdata}
\begin{figure}
\begin{tabular}{llll}
{\color[HTML]{000000} \begin{tabular}{c}
    \textbf{Original Image}  
\end{tabular}
} & {\ \begin{tabular}{c}
    \textbf{Drawn Bounding Box}
\end{tabular}
}  & { \begin{tabular}{c}
    \textbf{Ground Truth Mask}
\end{tabular}} \\
\includegraphics[width=3.6cm]{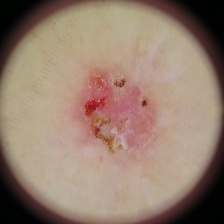}&
\includegraphics[width=3.6cm]{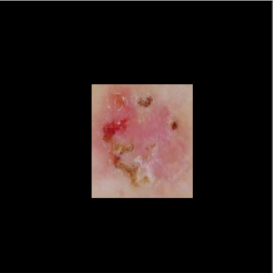}&
\includegraphics[width=3.6cm]{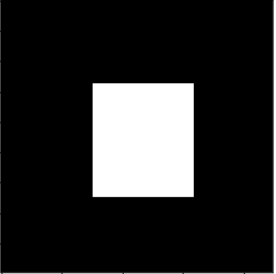}
\end{tabular}
 \caption{Ground truth labeling using the Bounding box technique}
  \label{boundbox}

\end{figure}

Task 1 data consists of about 2594 RGB images and their respective ground truths. The input images were resized to $224 \times 224$ with bicubic interpolation and normalized to the $[0,1]$ range. In order to check the classwise performance of our segmentation algorithm, we customly curated a segmentation dataset from some of the training images from the classification training dataset discussed in section \ref{classdata}. Segmentation ground truths were formed by drawing bounding boxs as depicted in fig. \ref{boundbox}.

In order to improve the classwise performance of the segmentation algorithm, the training set was augmented by populating it with more images and their respective ground truths drawn using the bounding box techniques. These images were chosen from the classification training set.
\subsection{Evaluation Metrics} \label{Metrics}
We have employed commonly used metrics such as Dice coefficient(DI) and Jaccard index (JI) to quantify image segmentation efficiency. Both these metrics essentially measure the similarity between the ground truth and the predicted segmented image in terms of the extent of overlap between the two images. 

Uncertainty is measured using monte carlo dropout and test time data augmentation. As mentioned in section \ref{uncertainity}, we calculate uncertainty $\varphi$ but the range of these values would change vary depending on the number of Monte Carlo samples. Hence we calculate normalised uncertainty $\varphi_{\text {norm}}$ where $\varphi_{\text {norm}}$ $\in[0,1]$.
\begin{center}
\begin{equation}
\scalebox{1.1}{
    $\varphi_{\text {norm}}=\frac{\varphi-\varphi_{\text {min}}}{\varphi_{\text {max}}-\varphi_{\text {min}}}$
    }
\end{equation}
\end{center}

To split the predictions into certain and uncertain categoris, we set a threshold $\varphi_{\text {T}}$ $\in[0,1]$ where a prediction is certain if $\varphi_{\text {norm}} < \varphi_{\text {T}}$ and uncertain if $\varphi_{\text {norm}} > \varphi_{\text {T}}$. 

When it comes to classification, we usually end up with 4 kinds of predictions i.e incorrect-uncertain \textit{(iu)},
correct-uncertain \textit{(cu)}, correct-certain \textit{(cc)}, and incorrect-certain \textit{(ic)} predictions. The overall accuracy of the uncertainty estimation could be expressed as a ratio of all the desirable cases i.e correct-certain \textit{(cc)} and incorrect-uncertain \textit{(iu)}, and all the possible cases. This diagnostic accuracy(A) can be represented in the form
\begin{center}
\begin{equation}
\scalebox{1}{
    $\operatorname{A}\left(\varphi_{T}\right)=\frac{L_{c c}+L_{i u}}{L_{c c}+L_{i u}+L_{c u}+L_{i c}}$
    }
\end{equation}
\end{center}
where $L$ represents the count for each possible combination.

In section \ref{explainability} different explainability techniques like GradCam, Guided Backprop, Guided GradCam , and XRAI have been discussed. In order to compare the performance of these techniques we have used bokeh effect and measured the accuracies as mentioned in \cite{kapishnikov2019xrai}. The basic intuition behind this analysis is that if the above explainability techniques identify important pixels to the model’s prediction then the model’s output of original image and reconstructed image must go hand in hand \cite{kapishnikov2019xrai}. Bokeh effect is used to reconstruct the image, in which initially the original image is blurred and the important pixels given by the explainability techniques are added. The explainability techniques used are thus compared using the prediction accuracy of the classification algorithm on these reconstructed images.
\subsection{Experimental Setup}
For the purpose of experimentation we have made use of a cloud based Nvidia RTX 2080ti GPU. The segmentation models i.e the U-Net and Bayesian MultiResUNet were trained with a learning rate of $10^{-3}$ and a batch size of 16. Different dropouts ranging between [0.4,0.7] were applied to get the best model which would not overfit on the training data and produce uncertainty estimates. A dropout rate of $0.5$ was found to be optimum. For classification, Bayesian DenseNet-169 and Bayesian ResNet50 models with dropouts were trained with a learning rate of $10^{-3}$ and a batch size of 16 and 32 respectively. Both the classification and segmentation models were trained using ADAM optimizer and binary cross entropy loss function Equation(\ref{bce}).

\begin{equation}\label{bce}
    H_{p}\left ( q \right ) = -  \frac{1}{C} \sum_{i=1}^{C}y_{i} \log(p(y_{i})) + (1-y_{i})\log(1-p(y_{i})) 
\end{equation}

\section{Results and Discussion}\label{resultsss}
In this section, we sequentially analyse the different stages of our SkiNet pipeline and prove its superiority. In section \ref{segresults}, we first compare the performance of various segmentation algorithms and justify why the Bayesian MultiResUnet is suitable for the first stage of the SkiNet pipeline. Followed by section \ref{classifresults}, where we discuss the performance of various state of the art classification algorithms and justify the usage of the Bayesian DenseNet-169 for the second stage of our pipeline. We also show prove that XRAI is the best explainability technique that could be used to explain the result to a diagnostician. Finally in section \ref{SKINETresults}, we discuss the performance of the SkiNet pipeline as a whole and prove the superiority of the SkiNet pipeline over the standalone DenseNet-169.
\subsection{Segmentation}\label{segresults}
In this section we analyse the performance of our different approaches towards the first step of our two stage pipeline. We mainly compare the performance of the U-Net and Bayesian MultiresUnet on the segmentation data. The class wise performance of our segmentation algorithm has also been assessed.

\begin{table}[H]
\caption{\normalsize{Comparitive study of various Segmentation models on the ISIC 2018 dataset\label{Segmenationresults}}}
\begin{tabular}{lll}
\rowcolor[HTML]{9B9B9B} 
{\color[HTML]{000000} \textbf{Model}}         & {\color[HTML]{000000} \textbf{DI}} & {\color[HTML]{000000} \textbf{JI}} \\
\rowcolor[HTML]{C0C0C0} 
{\color[HTML]{000000} U-Net}                  & {\color[HTML]{000000} 0.813}       & {\color[HTML]{000000} 0.734}       \\
\rowcolor[HTML]{9B9B9B} 
{\color[HTML]{000000} Bayesian MultiResUNet} & {\color[HTML]{000000} 0.852}       & {\color[HTML]{000000} 0.767}      
\end{tabular}
\end{table}

As observed in Table \ref{Segmenationresults}, the Bayesian MultiResUNet substantially outperforms the U-net on the test data. This could be attributed to the presence of the MultiRes blocks that considerably improves the detection of edges of the skin lesion.  The presence of multiple filter sizes in the MultiRes block could better explain this as it allows the Bayesain MultiResUNet to perform a pixel perfect segmentation.

\begin{figure}[H]
 \begin{tabular}{cccc}
  {\color[HTML]{000000} \begin{tabular}{c}
    \textbf{Original Image}  
\end{tabular}
} &  {\color[HTML]{000000} \begin{tabular}{c}
    \textbf{Ground Truth Mask}  
\end{tabular}
} &  {\color[HTML]{000000} \begin{tabular}{c}
    \textbf{U-Net Mask}  
\end{tabular}
} &
{\color[HTML]{000000} \begin{tabular}{c}
    \textbf{Bayesian}  \\
    \textbf{MultiResUNet} \\
    \textbf{Mask}
\end{tabular}
}
\\ \\
  \includegraphics[width=3cm]{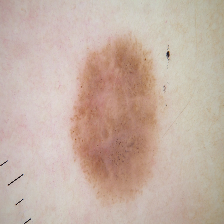}\label{ia} &   \includegraphics[width=3cm]{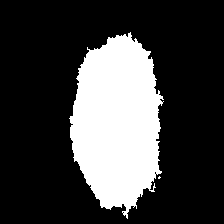}\label{ib} &
 \includegraphics[width=3cm]{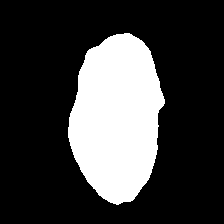}\label{ic} &   \includegraphics[width=3cm]{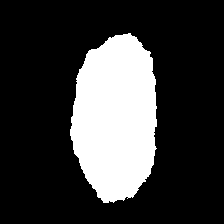}\label{ib} \\
(a) & (b)  &(c)  & (d) \\  
 \includegraphics[width=3cm]{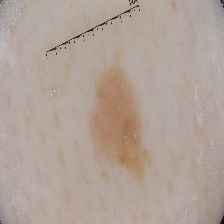}\label{ia} &   \includegraphics[width=3cm]{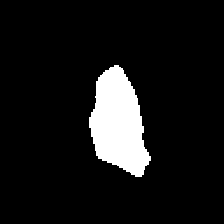}\label{ib} &
 \includegraphics[width=3cm]{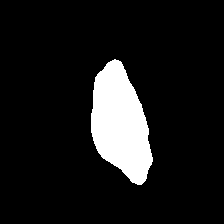}\label{ic} &   \includegraphics[width=3cm]{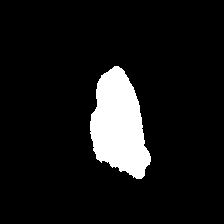}\label{ib} \\
(e) & (f)  &(g)  & (h) \\
\end{tabular}
\caption{Segmentation Results }\label{Fig:all_classes}
\end{figure}

In Fig. \ref{Fig:all_classes}, we clearly observe the Bayesian MultiResUNet outperform the U-Net with far more precise boundaries. The boundary produced by the MultiResUNet as observed in the mask Fig. \ref{Fig:all_classes}h is more accurate than the one produced by the U-Net as observed in Fig. \ref{Fig:all_classes}g, when we compare it to the ground truth mask in Fig. \ref{Fig:all_classes}f.  

\begin{table}[H]
\caption{\normalsize{Class wise performance of the Segmentation Algorithm\label{classiou}}}
\begin{tabular}{lllllllll}
\rowcolor[HTML]{9B9B9B} 
{\color[HTML]{000000} \textbf{Class}}                     & {\color[HTML]{000000} \textbf{MEL}} & {\color[HTML]{000000} \textbf{NV}} & {\color[HTML]{000000} \textbf{BCC}} & {\color[HTML]{000000} \textbf{AK}} & {\color[HTML]{000000} \textbf{BKL}} & {\color[HTML]{000000} \textbf{DF}} & {\color[HTML]{000000} \textbf{VASC}} & {\color[HTML]{000000} \textbf{SCC}} \\
\rowcolor[HTML]{C0C0C0} 
{\color[HTML]{000000} \textbf{JI (Before Augmentation)}} & {\color[HTML]{000000} 0.77}         & {\color[HTML]{000000} 0.85}        & {\color[HTML]{000000} 0.627}        & {\color[HTML]{000000} 0.64}        & {\color[HTML]{000000} 0.71}         & {\color[HTML]{000000} 0.751}       & {\color[HTML]{000000} 0.637}         & {\color[HTML]{000000} 0.677}        \\
\rowcolor[HTML]{9B9B9B} 
{\color[HTML]{000000} \textbf{JI (After Augmentation)}}  & {\color[HTML]{000000} 0.78}         & {\color[HTML]{000000} 0.855}       & {\color[HTML]{000000} 0.639}        & {\color[HTML]{000000} 0.65}        & {\color[HTML]{000000} 0.73}         & {\color[HTML]{000000} 0.78}        & {\color[HTML]{000000} 0.68}          & {\color[HTML]{000000} 0.72}        
\end{tabular}
\end{table}

\begin{figure}[H]
  \begin{tabular}{llll}

  {\color[HTML]{000000} \begin{tabular}{c}
    \textbf{Original Image}  
\end{tabular}
} &  {\color[HTML]{000000} \begin{tabular}{c}
    \textbf{Before Augmentation}  
\end{tabular}
} &  {\color[HTML]{000000} \begin{tabular}{c}
    \textbf{After Augmentation}  
\end{tabular}
} \\

    \\ { \subfloat[]{
 \includegraphics[width=3.6cm]{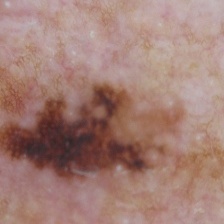}\label{ia}    \label{h}
}}   & { \subfloat[]{
\includegraphics[width=3.6cm]{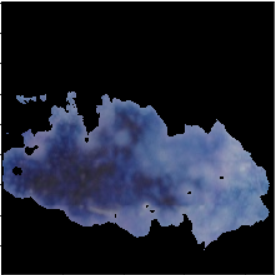}\label{ic}}} & 
{ \subfloat[]{
  \includegraphics[width=3.6cm]{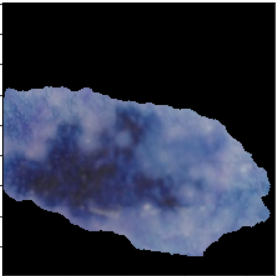}\label{ic}       
    \label{h}
}} \\
{ \subfloat[]{
  \includegraphics[width=3.6cm]{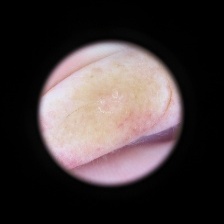}\label{ia}   
    \label{h}
}}

    &    
{ \subfloat[]{
  \includegraphics[width=3.6cm]{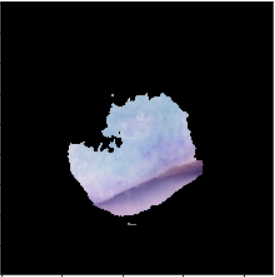}\label{ic}       
    \label{h}
}} &

     { \subfloat[]{
\includegraphics[width=3.6cm]{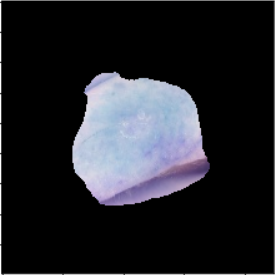}\label{ib}}}

\end{tabular}
\caption{Comparison of the Segmentation algorithm before and after data augmentation}
\label{segclassex}
\end{figure}

In order to confirm the suitability of our segmentation model as the first step of our two step pipeline we test it's performance on some of the classification. As observed in Table \ref{classiou}, the segmentation does perform well on some classes but not up to the mark on the others. In order to balance it out the dataset was augmented as discussed in section \ref{segdata} and the model was retrained. The average JI increased from 0.707 to about 0.729 and their has also been a class wise improvement which can be observed in Table \ref{classiou}. This data augmentation has not only helped in increasing the JI but has also helped in preserving much needed ROI that could help in producing a more accurate prediction in the next step i.e the classification step. In Fig. \ref{segclassex}b and \ref{segclassex}e we see certain regions that could help the classification algorithm being cropped by the segmentation algorithm. In Fig. \ref{segclassex}f the algorithm selects most of the brownish region without cropping any of it out when compared to the mask in Fig. \ref{segclassex}e. It also selects less of the unnecessary region when compared to Fig. \ref{segclassex}e.

We calculate the model uncertainty i.e, the epistemic uncertainty, aleatoric uncertainty as well as a combined uncertainty estimate that takes into account both epistemic and aleatoric uncertainty. We consider combined estimate for use in our SkiNet pipeline and set the uncertainty threshold $\varphi_{\text {T}}$ discussed in section \ref{Metrics} to be 0.25 when it comes to segmentation.

\begin{figure}[H]
\caption{\normalsize{Uncertainty Estimation of the Bayesian MultiResUNet\label{UncertainSegmenationresultsclass}}}

\begin{tabular}{llll}
{\color[HTML]{000000} \begin{tabular}{c}
    \textbf{Original Image}  
\end{tabular}
} & {\ \begin{tabular}{c}
    \textbf{Uncertainty Map} \\
    \textbf{(Aleatoric)}
\end{tabular}
}  & { \begin{tabular}{c}
    \textbf{Uncertainty Map}\\
    \textbf{(Epistemic)}
\end{tabular}}
&
{ \begin{tabular}{c}
    \textbf{Uncertainty Map} \\
    \textbf{(Aleatoric + Epistemic)}
\end{tabular}} \\

 { \subfloat[]{
     \includegraphics[width=2.4cm]{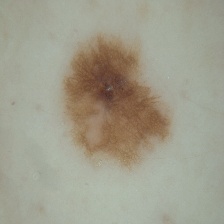}
    \label{h}
}} & {\subfloat[$\varphi_{norm} = 0.169$]{
     \includegraphics[width=2.4cm]{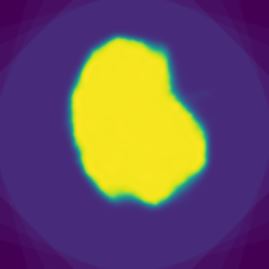}
    \label{h}
}} & { \subfloat[$\varphi_{norm} = 0.239$]{
     \includegraphics[width=2.4cm]{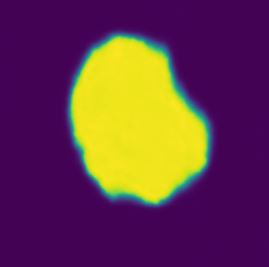}
    \label{h}
}}& { \subfloat[$\varphi_{norm} = 0.176$]{
     \includegraphics[width=2.4cm]{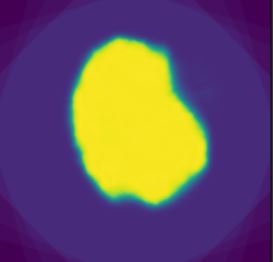}
    \label{h}
}}   \\ { \subfloat[]{
     \includegraphics[width=2.4cm]{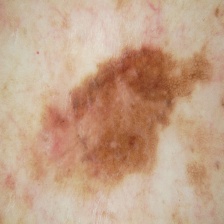}
    \label{h}
}} &{ \subfloat[$\varphi_{norm} = 0.452$]{
     \includegraphics[width=2.4cm]{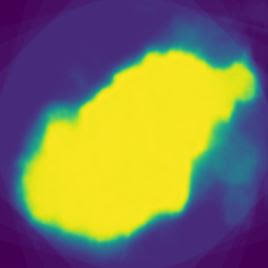}
    \label{h}
}} & { \subfloat[$\varphi_{norm} = 0.456$]{
     \includegraphics[width=2.4cm]{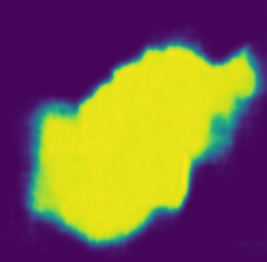}
    \label{h}
}} & {\subfloat[$\varphi_{norm} = 0.447$]{
     \includegraphics[width=2.4cm]{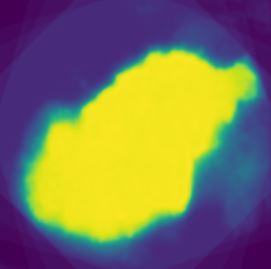})
    \label{h}
}}\\ { \subfloat[]{
     \includegraphics[width=2.4cm]{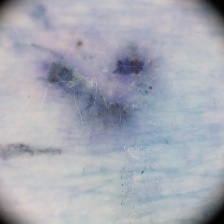}
    \label{h}
}}
    & { \subfloat[$\varphi_{norm} = 0.491$]{
     \includegraphics[width=2.4cm]{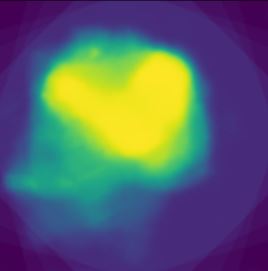}
    \label{h}
}}
    & { \subfloat[$\varphi_{norm} = 0.580$]{
     \includegraphics[width=2.4cm]{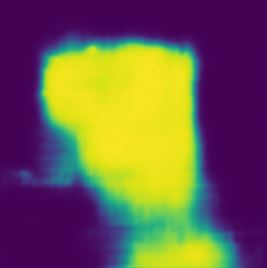}
    \label{h}
}}
     & { \subfloat[$\varphi_{norm} = 0.491$]{
     \includegraphics[width=2.4cm]{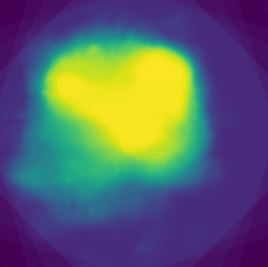}
    \label{h}
}}   \\

\end{tabular}
\end{figure}

As observed in Fig. \ref{UncertainSegmenationresultsclass}, the light greenish blue region represents the uncertain region in our segmentation map. From Fig \ref{UncertainSegmenationresultsclass}, we observe that the segmentation map produced for image (a) is certain as its uncertainty is well within the defined threshold and its uncertain region is negligible. The same cannot be said regarding the maps produced for images Fig \ref{UncertainSegmenationresultsclass}(b) and Fig \ref{UncertainSegmenationresultsclass}(c). We do observe that the combined uncertainty is higher than the defined threshold and that a significant region is highlighted as uncertain in both maps. We could also observe that the combined uncertainty maps look quite similar to that of the aleatoric uncertainty maps and that the Combined uncertainty score is close to the aleatoric uncertainty score. Hence we could say that the Aleartoric uncertainty has a greater contibution to the overall uncertainty.

 \begin{figure}[!ht]
\begin{tabular}{cccc}
  \includegraphics[width=2.4cm]{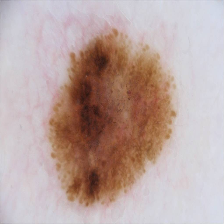}\label{ia} &   \includegraphics[width=2.4cm]{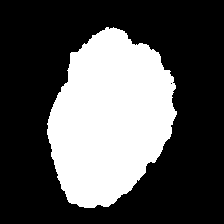}\label{ib} &
 \includegraphics[width=2.4cm]{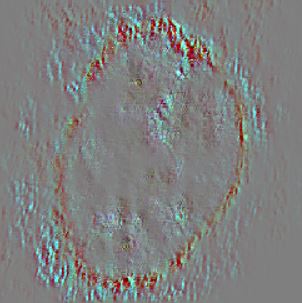}\label{ic} &   \includegraphics[width=2.4cm]{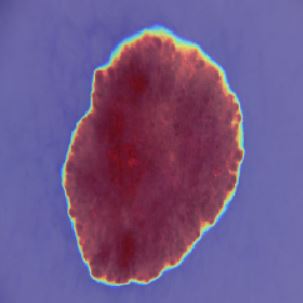}\label{ib} \\
(a) & (b)  &(c)  & (d)  
\end{tabular}
\caption{Interpretability of the Bayesian MultiResUNet (a)Original Image (b)Ground Truth (c)Guided Backprop (d)Grad CAM Heatmap}
\label{segexplain}
\end{figure}

Fig.  \ref{segexplain}(c) and \ref{segexplain}(d) show examples of the use of  Guided Backprop and Grad CAM in order to visualise the Bayesian MultResUNet. We do observe that the model takes edges into consideration in order to produce the segmentation mask. This could be observed in Fig. \ref{segexplain}(d) where the important pixels are highlighted in yellow which signify the boundary of the image. The same can be said about the interpretation using Guided Backprop too, as represented in Fig. \ref{segexplain} (c) where the important region is highlighted with reddish-yellowish pixels and less important region is highlighted using light greenish pixels.

\subsection{Classification}\label{classifresults}
In this section, we analyse the performance of various off the shelf state of the art architectures against the ISIC 2018 task 3 dataset. We map the uncertainty for every prediction made and classify each prediction as certain or uncertain. Explainability techniques discussed in section 3.2 have been used to visualize the learning of the classication algorithm and have thus been compared to equip our SkiNet pipeline with the best possible explainability technique.

\begin{table}[H]

\caption{\normalsize{Comparitive study of various Classification models on the ISIC 2018 dataset\label{classificationresults}}}
\scalebox{1.2}{
\begin{tabular}{ll}
\rowcolor[HTML]{9B9B9B} 
{\color[HTML]{000000} \textbf{Model}}     & {\color[HTML]{000000} \textbf{Prediction Accuracy(\%)}} \\
\rowcolor[HTML]{C0C0C0} 
{\color[HTML]{000000} ResNet-50}          & {\color[HTML]{000000} 84.87}                            \\
\rowcolor[HTML]{9B9B9B} 
{\color[HTML]{000000} Bayesian ResNet-50} & {\color[HTML]{000000} 85.13}                            \\

\rowcolor[HTML]{C0C0C0} 
DenseNet-169                              & 86.67                                                   \\
\rowcolor[HTML]{9B9B9B} 
Bayesian DenseNet-169                     & 87.35                                                                                       
\end{tabular}
}
\end{table}

From Table \ref{classificationresults} we observe that the bayesian models ResNet-50 and DenseNet-169 outperform the standard versions of these architectures. The addition of Monte Carlo dropout has helped in boosting the performance of these standard models. Due to the superior performance of the Bayesian DenseNet-169 over the Bayesian ResNet-50, we employ the use of the Bayesian DenseNet-169 for the second step in our pipeline. Our bayesian model outputs posterior probability distributions over each of the seven classes.
\begin{table}[H]
\caption{\normalsize{Uncertainty of the Bayesian DenseNet-169\label{classificationuncertain}}}
\begin{tabular}{lll}
\rowcolor[HTML]{9B9B9B} 
{\color[HTML]{000000} \textbf{Epsitemic}} & {\color[HTML]{000000} \textbf{Aleatoric}} & {\color[HTML]{000000} \textbf{Combined(Epsitemic+Aleatoric)}} \\
\rowcolor[HTML]{C0C0C0} 
0.102                                      & 0.269                                     & 0.283                                                        
\end{tabular}
\end{table}

\begin{figure}[H]
    
    \subfloat[Correct-Certain]{
\begin{tabular}{cc}
  \includegraphics[width=3cm]{img0.JPG}\label{ia1} &   \includegraphics[width=3cm]{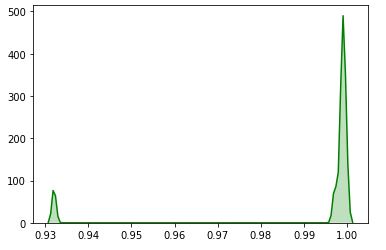}\label{ib1} \\
(i) Original Image & (ii) Posterior Distribution\\

              & $\varphi_{\text {norm}} = 0.13 $ 

     \\[6pt]
 
\end{tabular}
\label{cc}
    }
    \subfloat[Incorrect-Certain]{
\begin{tabular}{cc}
  \includegraphics[width=3cm]{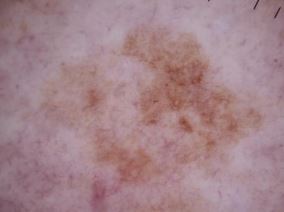}\label{ia2} &   \includegraphics[width=3cm]{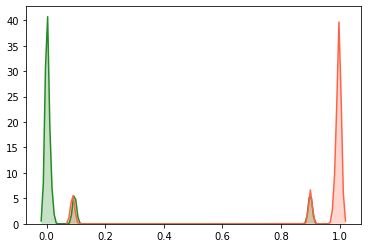}\label{ib2} \\
(i) Original Image & (ii) Posterior Distribution\\

              & $\varphi_{\text {norm}} = 0.18 $  \\[6pt]

\end{tabular}

    }
    
    \hspace{2mm}
    
    \subfloat[Correct-Uncertain]{
\begin{tabular}{cc}
  \includegraphics[width=3cm]{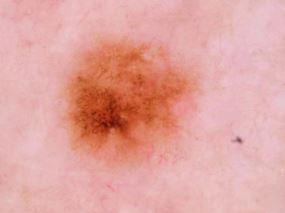}\label{ia3} &   \includegraphics[width=3.6cm]{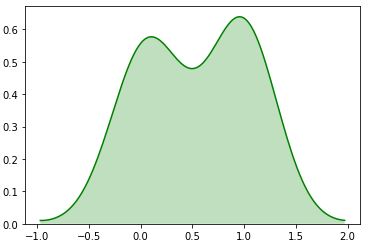}\label{ib3} \\
(a) Original Image & (b) Posterior Distribution\\

              & $\varphi_{\text {norm}} = 0.602 $
\\[6pt]
 
\end{tabular}
    }
    \subfloat[Incorrect-Uncertain]{
\begin{tabular}{cc}
  \includegraphics[width=3cm]{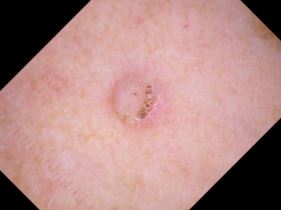}\label{ia4} &   \includegraphics[width=3cm]{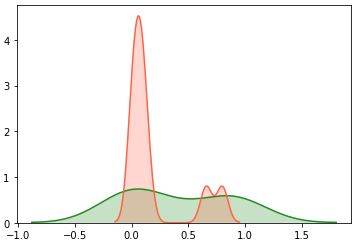}\label{ib4} \\
(a) Original Image & (b) Posterior Distribution\\

              & $\varphi_{\text {norm}} = 0.76$ \\[6pt]
\end{tabular}

}

\caption{Posterior Probability Distributions for each of the possible scenarios i.e incorrect-uncertain \textit{(iu)},
correct-uncertain \textit{(cu)}, correct-certain \textit{(cc)}, and incorrect-certain \textit{(ic)} predictions assuming that the combined $\varphi_T$ is 0.35. The red region indicates the  posterior probability distribution for the incorrect class where as the green region indicates the posterior probability distribution of the correct class.}\label{Fig:class_uncertain}
\end{figure}

From Table \ref{classificationuncertain} we observe that the aleatoric uncertainty is higher when compared to the epistemic uncertainty and shows to have a higher influence on the overall combined uncertainty. This could be attributed to severe class imbalance of the ISIC 2018 dataset and the augmentations of the image data which have been done as discussed in section \ref{datasets} might be a contributing factor. Examples of posterior probability distributions for each category discussed in section \ref{Metrics} could be observed in Fig. \ref{Fig:class_uncertain}. Interpretability techniques like Guided Grad CAM and XRAI have been used to depict the region of interest in predcitions of the correct-certain \textit{(cc)} category.

\begin{figure}[H]
\caption{Model's Region of Interest depicted by XRAI and Guided Grad CAM }

\begin{tabular}{llll}
{\color[HTML]{000000} \begin{tabular}{c}
    \textbf{Original Image}  
\end{tabular}
} & { \begin{tabular}{c}
    \textbf{XRAI Heatmap}
\end{tabular}
}  & { \begin{tabular}{c}
    \textbf{Guided GradCam}
\end{tabular}}
\\
\subfloat[]{
  \includegraphics[width=3.6cm]{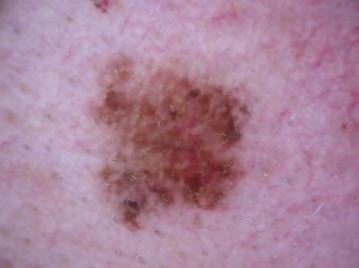}
  \label{eed}
} &
\subfloat[]{
  \includegraphics[width=3.6cm]{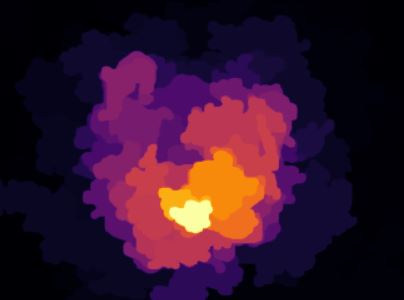}
  \label{eee}
} &
\subfloat[]{
  \includegraphics[width=3.6cm]{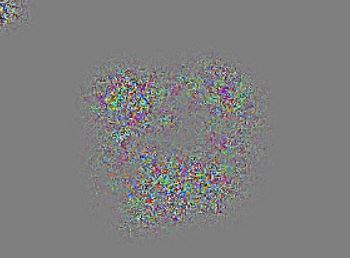}
  \label{eef}
  }
  \\
     \subfloat[]{
  \includegraphics[width=3.6cm]{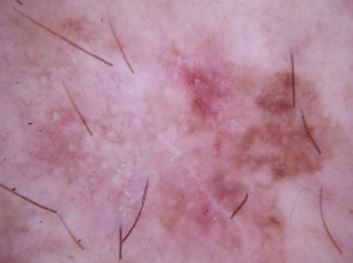}
  \label{eeg}
} &
\subfloat[]{
  \includegraphics[width=3.6cm]{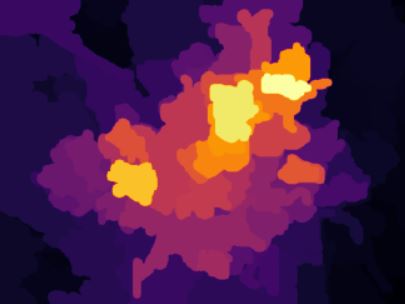}
  \label{eeh}
} &
\subfloat[]{
  \includegraphics[width=3.6cm]{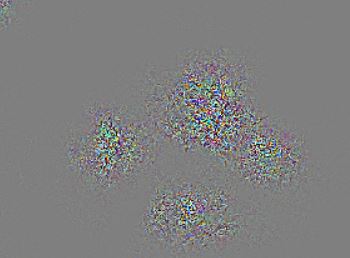}
  \label{eei}
  
}
\end{tabular}
\label{vismap}
\end{figure}

\begin{figure}[H]

\begin{tabular}{llll}
  
{\color[HTML]{000000} \begin{tabular}{c}
    \textbf{Original Image}  
\end{tabular}
} & {\ \begin{tabular}{c}
    \textbf{Top 20\%}
\end{tabular}
}  & { \begin{tabular}{c}
    \textbf{Top 10\%}
\end{tabular}}
&
{ \begin{tabular}{c}
    \textbf{Top 5\%}
\end{tabular}}
  \\
  \subfloat[]{
  \includegraphics[width=3cm]{img3.JPG}\label{e1ia}} &   \subfloat[]{\includegraphics[width=3cm]{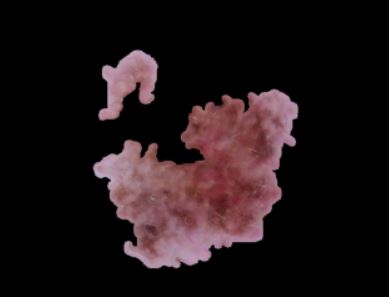}\label{e1ib} }&
\subfloat[]{
 \includegraphics[width=3cm]{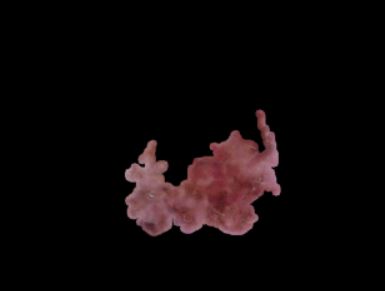}\label{e1ic}} &  
 \subfloat[]{\includegraphics[width=3cm]{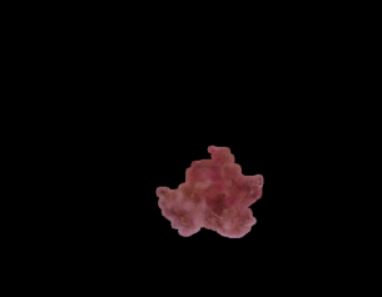}\label{e1id}} \\

  \subfloat[]{\includegraphics[width=3cm]{img4.JPG}\label{e2ie}} &   \subfloat[]{\includegraphics[width=3cm]{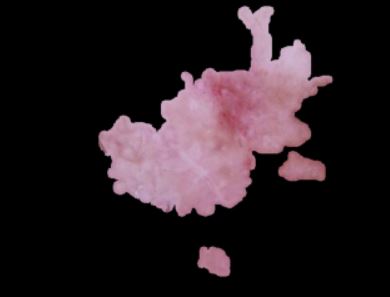}\label{e2if}} &
 \subfloat[]{
 \includegraphics[width=3cm]{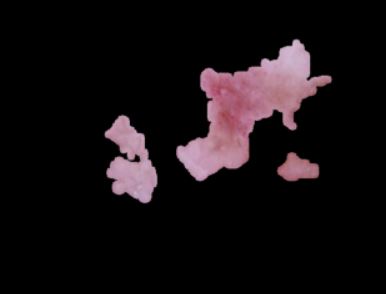}\label{e2ig}} & 
 \subfloat[]{\includegraphics[width=3cm]{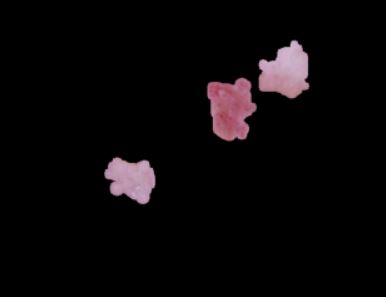}\label{e2ih}} \\

\end{tabular}
    \caption{Top regions of Interest identified by XRAI}\label{xraiin}
\end{figure}

For the skin lesion in Fig. \ref{vismap}a we observe that the bottom right part of the skin lesion depicted in the XRAI heatmap in Fig.  \ref{vismap}b and Guided Grad CAM map in Fig.  \ref{vismap}c is of importance to our model. This is clearly depicted by the top 10 \% and top 5 \% plots in Figs. \ref{e1ic} and \ref{e1id}, showing that the model was heavily influenced by the dark red region. From Fig. \ref{vismap}e, we observe that the top right part of the lesion is of importance to the model. Figs. \ref{e2ig} and \ref{e2ih} show that the model is influenced by the reddish pinkish region present in the top right part of the skin lesion. 


\begin{table}[H]
\caption{Comparitive Analysis between different Explainability Techniques}
\label{ExpComp}
\begin{tabular}{ll}
\rowcolor[HTML]{9B9B9B} 
{\color[HTML]{000000} \textbf{Explainability Technique}} & {\color[HTML]{000000} \textbf{Accuracy}} \\
\rowcolor[HTML]{C0C0C0} 
{\color[HTML]{000000} GradCam}                           & {\color[HTML]{000000} 73\%}              \\
\rowcolor[HTML]{9B9B9B} 
{\color[HTML]{000000} Guided Backprop}                   & {\color[HTML]{000000} 73\%}              \\
\rowcolor[HTML]{C0C0C0} 
{\color[HTML]{000000} Guided GradCam}                    & {\color[HTML]{000000} 77\%}              \\
\rowcolor[HTML]{9B9B9B} 
{\color[HTML]{000000} XRAI}                              & {\color[HTML]{000000} 84\%}             
\end{tabular}
\end{table}

The explainability techniques have been compared as discussed in section \ref{Metrics} and from Table \ref{ExpComp} we can clearly say that XRAI provides a more clear visualisation of what our classification algorithm is learning. Hence we conclude that XRAI would be the best fit for the SKINET pipeline and thus provide the best possible explanation behind the prediction.
\subsection{SkiNet Performance}\label{SKINETresults}
In this section, we compare the performance of our SkiNet pipeline compared to that of a stand-alone Bayesian DenseNet-169. We also showcase how a certain mask produced by the Bayesian MultiResUNet in the first step of the SkiNet pipeline helps in reducing uncertainty and increase the overall accuracy.
\begin{table}[H]
\caption{\normalsize{Categorical Segregation of predictions made on our test data\label{testuncertaincomp}}}
\begin{tabular}{lll}
\rowcolor[HTML]{9B9B9B} 
{\color[HTML]{000000} \textbf{Category}}       & {\color[HTML]{000000} \textbf{Stand-alone Bayesian DenseNet-169}} & {\color[HTML]{000000} \textbf{SkiNet Pipeline}} \\
\rowcolor[HTML]{C0C0C0} 
{\color[HTML]{000000} Correct Certain(cc)}     & {\color[HTML]{000000} 1602}                                       & {\color[HTML]{000000} 1727}                     \\
\rowcolor[HTML]{9B9B9B} 
{\color[HTML]{000000} Correct Uncertain(uc)}   & {\color[HTML]{000000} 722}                                        & {\color[HTML]{000000} 627}                      \\
\rowcolor[HTML]{C0C0C0} 
{\color[HTML]{000000} Incorrect Certain(ic)}   & {\color[HTML]{000000} 76}                                         & {\color[HTML]{000000} 74}                       \\
\rowcolor[HTML]{9B9B9B} 
{\color[HTML]{000000} Incorrect Uncertain(iu)} & {\color[HTML]{000000} 261}                                        & {\color[HTML]{000000} 233}                     
\end{tabular}
\end{table}
From Table \ref{testuncertaincomp}, we clearly observe that SkiNet has a better overall diagnostic accuracy(A) of 73.65\% when compared to the 70.01\% of the stand-alone Bayesian DenseNet-169. It also performs better in terms of prediction accuracy with an accuracy of 88.46\% when compared to the 87.35\% of the Bayesian DenseNet-169 as seen in Table \ref{classificationresults}.
 

\begin{figure}[H]
    \subfloat[Stand-Alone DenseNet-16]{
     \includegraphics[width=6cm]{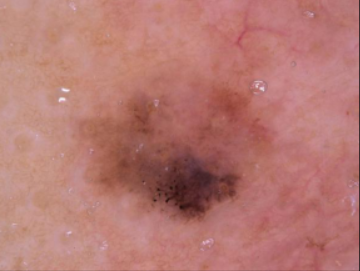}

    \label{h}

}                                                                                    
\subfloat[SkiNet Pipeline]{

     \includegraphics[width=6cm]{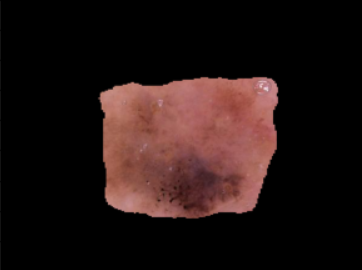}

    \label{h}

}
    \caption{Image received by the Classification Algorithm}
    \label{Fig:280asbs}
\end{figure}

\begin{table}[H]
\caption{Comparitive Analysis Between the performance of a Stand-alone DenseNet-169 and the SkiNet Pipeline [CU \textrightarrow CC]}\label{CUCC}
\begin{tabular}{l
>{\columncolor[HTML]{C0C0C0}}l 
>{\columncolor[HTML]{C0C0C0}}l }
                                                                                                                                                 & \cellcolor[HTML]{9B9B9B}{\color[HTML]{000000} \textbf{Stand-Alone DenseNet-169}} & \cellcolor[HTML]{9B9B9B}{\color[HTML]{000000} \textbf{SkiNet Pipeline}} \\
\cellcolor[HTML]{9B9B9B}{\color[HTML]{000000} \textbf{Ground Truth}}                                                                             & MEL                                                                              & MEL                                                                     \\
\cellcolor[HTML]{9B9B9B}{\color[HTML]{000000} \textbf{Prediction}}                                                                               & MEL                                                                              & MEL                                                                     \\
\cellcolor[HTML]{9B9B9B}{\color[HTML]{000000} \textbf{Uncertainty}}                                                                              & 0.68                                                                             & 0.30                                                                    \\
\cellcolor[HTML]{9B9B9B}{\color[HTML]{000000} \textbf{Category}}                                                                                 & Correct Uncertain                                                                               & Correct Certain    \\

\end{tabular} 
 \\

\end{table}

\begin{figure}[H]
    \centering
  \subfloat[Before Segmentation]{
        \includegraphics[width=6cm]{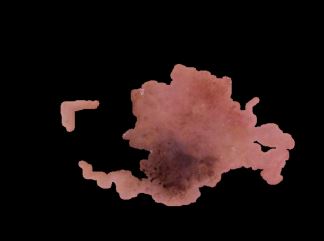}
  }
    \subfloat[After Segmentation]{
        \includegraphics[width=6cm]{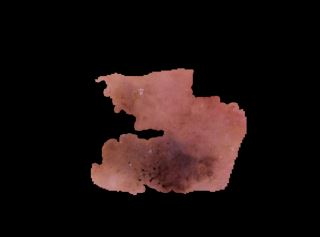}
  }
    \caption{Regions of Interest identified by XRAI a.) Before Segmentation b.) After Segmentation}
    \label{Fig:CUCCcomp}
\end{figure}
Sometimes, though a prediction maybe correct, it may be deemed as uncertain due to the high uncertainty which is mainly caused by the presence of noise in the image. In the case of dermoscopic images, it is mainly in the form of sweat droplets, hair, other lesions etc. This noise could be reduced with the use of segmentation which would crop the unnecessary part out and highlight the main region of the lesion. This phenomenon is clearly observed in Fig. \ref{Fig:280asbs}, where sweat droplets and the unnecessary background is cropped out by the Bayesian MultiResUNet present in the SkiNet pipeline. This improvement can be clearly observed in the XRAI map in Fig. \ref{Fig:CUCCcomp} where we clearly see that the algorithm now focuses on the tumor itself rather than the unnecessary background. From Table \ref{CUCC}, We observe the uncertainty drop from 0.68 to 0.30 which is within the set threshold $\varphi_T$ of 0.35 thus leading to a certain prediction.
\begin{figure}[H]
    \subfloat[Stand-Alone DenseNet-16]{
     \includegraphics[width=6cm]{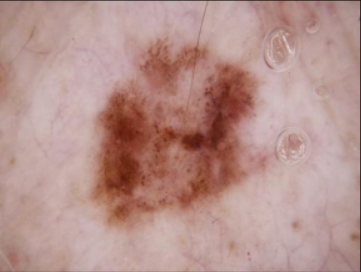}

    \label{h}

}                                                                                    
\subfloat[SkiNet Pipeline]{

     \includegraphics[width=6cm]{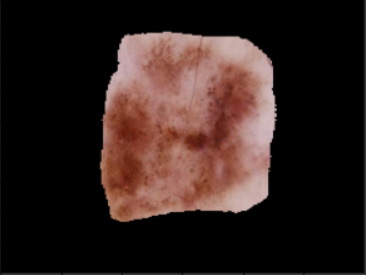}

    \label{h}

}
    \caption{Image received by the Classification Algorithm}
    \label{Fig:1120asbs}
\end{figure}
\begin{table}[H]
\caption{Comparitive Analysis Between the performance of a Stand-alone DenseNet-169 and the SkiNet Pipeline [IU \textrightarrow CC]}\label{IUCC}
\begin{tabular}{l
>{\columncolor[HTML]{C0C0C0}}l 
>{\columncolor[HTML]{C0C0C0}}l }
                                                                                                                                                 & \cellcolor[HTML]{9B9B9B}{\color[HTML]{000000} \textbf{Stand-Alone DenseNet-169}} & \cellcolor[HTML]{9B9B9B}{\color[HTML]{000000} \textbf{SkiNet Pipeline}} \\
\cellcolor[HTML]{9B9B9B}{\color[HTML]{000000} \textbf{Ground Truth}}                                                                             & MEL                                                                              & MEL                                                                     \\
\cellcolor[HTML]{9B9B9B}{\color[HTML]{000000} \textbf{Prediction}}                                                                               & NV                                                                              & MEL                                                                     \\
\cellcolor[HTML]{9B9B9B}{\color[HTML]{000000} \textbf{Uncertainty}}                                                                              & 0.45                                                                             & 0.04                                                                   \\
\cellcolor[HTML]{9B9B9B}{\color[HTML]{000000} \textbf{Category}}                                                                                 & Incorrect Uncertain                                                                             & Correct Certain    \\

\end{tabular} 
 \\

\end{table}
\begin{figure}[H]
    \centering
  \subfloat[Before Segmentation]{
        \includegraphics[width=6cm]{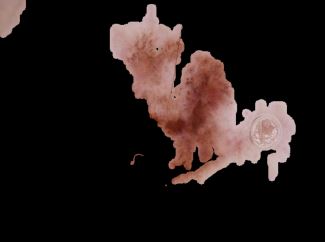}
  }
    \subfloat[After Segmentation]{
        \includegraphics[width=6cm]{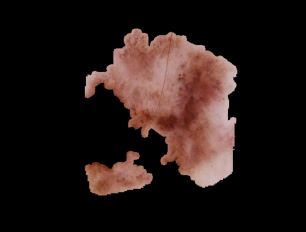}
  }
    \caption{Regions of Interest identified by XRAI a.) Before Segmentation b.) After Segmentation}
    \label{Fig:IUCCcomp}
\end{figure}
In Fig. \ref{Fig:IUCCcomp}a, we see that the classification algorithm is heavily influenced by a water droplets present in the background thus leading to a misclassification. In Fig. \ref{Fig:1120asbs}, we observe that Bayesian MultiResUNet does a good job in getting rid of the unnecessary background and highlighting the region of interest in the lesion. This can be observed in Fig. \ref{Fig:IUCCcomp}b where the classification algorithm concentrates on the different regions of the lesion. From Table \ref{IUCC} we clearly observe that this cropping out has helped the classification algorithm in making a correct and confident prediction. 
\begin{figure}[H]
    \subfloat[Stand-Alone DenseNet-16]{
     \includegraphics[width=6cm]{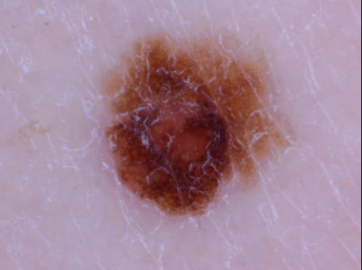}

    \label{h}

}                                                                                    
\subfloat[SkiNet Pipeline]{

     \includegraphics[width=6cm]{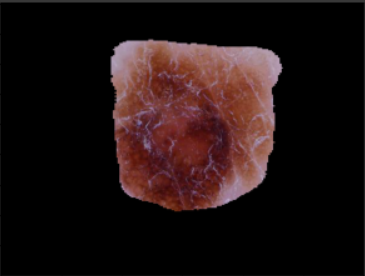}

    \label{h}

}
    \caption{Image received by the Classification Algorithm}
    \label{Fig:1718asbs}
\end{figure}

\begin{table}[H]
\caption{Comparitive Analysis Between the performance of a Stand-alone DenseNet-169 and the SkiNet Pipeline [IC \textrightarrow CC]}\label{ICCC}
\begin{tabular}{l
>{\columncolor[HTML]{C0C0C0}}l 
>{\columncolor[HTML]{C0C0C0}}l }
                                                                                                                                                 & \cellcolor[HTML]{9B9B9B}{\color[HTML]{000000} \textbf{Stand-Alone DenseNet-169}} & \cellcolor[HTML]{9B9B9B}{\color[HTML]{000000} \textbf{SkiNet Pipeline}} \\ 
\cellcolor[HTML]{9B9B9B}{\color[HTML]{000000} \textbf{Ground Truth}}                                                                             & MEL                                                                              & MEL                                                                     \\
\cellcolor[HTML]{9B9B9B}{\color[HTML]{000000} \textbf{Prediction}}                                                                               & NV                                                                              & MEL                                                                     \\
\cellcolor[HTML]{9B9B9B}{\color[HTML]{000000} \textbf{Uncertainty}}                                                                              & 0.34                                                                             & 0.12                                                                   \\
\cellcolor[HTML]{9B9B9B}{\color[HTML]{000000} \textbf{Category}}                                                                                 & Incorrect Certain                                                                             & Correct Certain    \\

\end{tabular} 
 \\

\end{table}

\begin{figure}[H]
    \centering
  \subfloat[Before Segmentation]{
        \includegraphics[width=6cm]{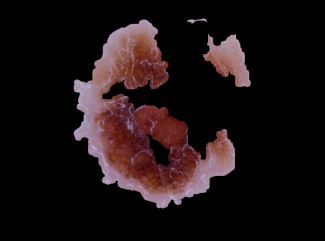}
  }
    \subfloat[After Segmentation]{
        \includegraphics[width=6cm]{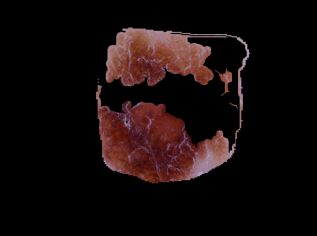}
  }
    \caption{Regions of Interest identified by XRAI a.) Before Segmentation b.) After Segmentation}
    \label{Fig:CUCCcomp}
\end{figure}

From Fig. \ref{Fig:1718asbs} we observe that the Bayesian MultiResUNet present in the first step of our pipeline helps in enhancing the lesion of the image and helps in cropping out the unnecessary background. In Fig. \ref{Fig:CUCCcomp} we observe that the area of interest for the classification algorithm rather remains similar before and after segmentation but with expulsion of the unnecessary background it becomes quite clear for the algorithm to make a prediction. Thus leading to an accurate certain prediction as suggested by Table \ref{CUCC}.

\section{Conclusion}\label{conc}
This article addresses the need to integrate explainability and uncertainty modeling in the automated skin lesion diagnosis process. In this paper, we have proposed a novel SkiNet pipeline for the diagnosis of skin lesion. The proposed Bayesian Multi ResUNet which is used for segmentation, also produces uncertainty maps to incorporate the confidence measure. The DenseNet-169 with added dropout has been used for classification and has demonstrated superior performance over the original. The addition of segmentation as a pre-processing step for classification has greatly impacted the efficiency of the classification model. The uncertainty score of the segmentation model’s output is used to pass only the most confident predictions to classification model. The uncertainty score of the classification model tests the confidence of the model ’s prediction and suggests second opinion in the event of less positive predictions thereby reducing misdiagnosis to some degree. The diagnostic accuracy of stand-alone Bayesian DenseNet-169 is 70.01\%, which further improved to 73.65\% after performing segmentation using the proposed SkiNet pipeline. When deploying such models, one could use model explanations to “gate” the use of the machine learning system. To build trust of the medical community in the proposed model, we use an explainability map that shows the salient region for the model. Using the saliency maps provided by various techniques such as GradCAM, Guided Backprop, Guided GradCAM and XRAI, the original images are reconstructed with the aid of Bokeh effect. They are then passed through the classification model and the accuracy scores thus obtained clearly demonstrate a superior performance of XRAI with an enhanced 84\% accuracy. The results of the proposed pipeline is quite encouraging and can be generalized for other similar tasks in the medical domain. This article has used post-hoc interpretability methods however, we would also like to explore some pre-hoc interpretation methods like attention mechanism while training the model in order to further enhance the model’s performance.


\bigskip
\textbf{Conflict of interest:} The authors declare that they have no conflicts of interest.
\bibliographystyle{spmpsci}      



\end{document}